# IDIOTYPIC IMMUNE NETWORKS IN MOBILE ROBOT CONTROL

Amanda M. Whitbrook, Uwe Aickelin, *Member, IEEE*, and Jonathan M. Garibaldi

*Abstract*—Jerne's idiotypic network theory postulates that the immune response involves inter-antibody stimulation and suppression as well as matching to antigens. The theory has proved the most popular Artificial Immune System (AIS) model for incorporation into behavior-based robotics but guidelines for implementing idiotypic selection are scarce. Furthermore, the direct effects of employing the technique have not been demonstrated in the form of a comparison with non-idiotypic systems. This paper aims to address these issues. A method for integrating an idiotypic AIS network with a Reinforcement Learning based control system (RL) is described and the mechanisms underlying antibody stimulation and suppression are explained in detail. Some hypotheses that account for the network advantage are put forward and tested using three systems with increasing idiotypic complexity. The basic RL, a simplified hybrid AIS-RL that implements idiotypic selection independently of derived concentration levels and a full hybrid AIS-RL scheme are examined. The test bed takes the form of a simulated Pioneer robot that is required to navigate through maze worlds detecting and tracking door markers.

## I. INTRODUCTION

THE main focus of mobile robot research has been behavior-based reactive control since the publication of Brooks' subsumption architecture in the mid-eighties [22]. This approach allows a degree of intelligence to emerge from competence module (individual behavior) interactions, but it is normally integrated with other

AI methods, for example reinforcement learning [17] or neural networks [14], as these provide greater flexibility for dynamically changing environments. More recently, researchers have been exploiting the learning and adaptive properties of the vertebrate immune system in order to design effective sensory response algorithms. In essence the immune system matches antibodies (receptors on b-cells) to antigens (foreign material that invades the body), so that b-cells with suitable receptors undergo stimulation, increase in number (clonal selection) and destroy the invading cells. Artificial Immune Systems (AIS) have a matching function that determines the strength of the bond between the antibody and antigen and they utilize a concentration parameter as an additional measure of antibody fitness. A comprehensive introduction to AIS systems and their applications is provided in [4].

Within mobile robotics, Farmer's [2] computational model of Jerne's idiotypic network theory [1] has been notable as a means of inducing flexible behavior mediation and it has demonstrated some encouraging results. In these idiotypic networks, antibodies (competence modules) are linked both to antigens (environmental stimuli) and to each other, forming a dynamic chain of suppression and stimulation that affects their concentration levels globally. The system is balanced so that concentration levels also play a role in determining the degrees of stimulation and suppression that occur. This "global perspective" differs from the more conventional AIS approach (clonal selection theory [3]), which considers that only antibody-antigen stimulation alters antibody concentrations.

The success of the idiotypic systems has largely been attributed to the behavior arbitration capabilities of the communicating antibodies, but no attention has been directed towards proving that this is the case, or showing that other systems are inferior. In addition, there has been little attempt to explain the particular mechanisms by which antibodies stimulate and suppress each other and how this is able to improve robot performance. This paper aims to address these issues by providing a comprehensive description of a hybrid robot control system that implements Reinforcement Learning (RL) with a Farmer-based idiotypic network for antibody selection. Although the system described does not attempt to evolve network connections and uses a fixed set of antibodies, additional details missing from earlier narratives are supplied. In particular, a rigorous account of the implementation of stimulation and suppression and some hypotheses that try to explain the idiotypic advantage are given. Most importantly, this paper seeks to test these hypotheses by undertaking a number of experiments that introduce idiotypic effects into the RL system gradually.

In the first system (S1) an idiotypic network is not implemented and antibodies are selected on the basis of strength of match to antigens only. In effect this is a pure reinforcement learning system. The second system S2 is a hybrid that couples reinforcement learning with a simplified idiotypic network. Antibodies are selected by summing the effects of the network interactions to provide a global strength of match, but concentration levels do not influence the idiotypic process in any way. The third system S3 is a full AIS that bases selection on a combination of the global strength of match and the concentration level and also feeds the concentrations levels back to the network. This step-wise approach is important in attempting to assess and explain the effects of introducing the idiotypic network into the system. In addition, idiotypic dynamics have not previously been uncoupled from antibody concentrations when implementing the Farmer equation, which represents a novel investigation.

The paper is arranged as follows. Section II provides background information including a brief account of the biological immune system that highlights the main differences between the more traditional clonal selection theory [3] and Jerne's idiotypic network theory [1]. The section also describes how the network theory has been applied to autonomous robot navigation and a short review of recent work in this field is given. Section III discusses the motivation behind the research, relating the problems associated with reinforcement learning and introducing some hypotheses that attempt to explain the idiotypic network advantage. Section IV details the navigation problem and environments that have been used as the test bed for the hypotheses. Section V presents information on system architecture, Section VI focuses on the experimental methodology adopted and Section VII reports on the results and their interpretation. Section VIII concludes the paper.

## II. BACKGROUND

### A. Clonal Selection Theory

In the adaptive immune system of vertebrates, b-cells play an important role in the identification and removal of antigens. The clonal selection theory [3] states that division occurs for b-cells with receptors that have a high degree of match to a stimulating antigen's epitope pattern and that these cells then mature into plasma cells that secrete the matching receptors or antibodies into the bloodstream. The reproduction of the b-cells also causes a high rate of mutation so that weakly matching cells may mutate to produce antibodies with higher affinities for the stimulating antigen. Once in the bloodstream the antibody combining sites or paratopes bind to the antigen epitopes, causing other cells to assist with elimination. Paratopes and epitopes are complimentary and are analogous to keys and locks. Paratopes can be viewed as master keys that may open a set of locks and some locks can be opened by more than one key [2].

Some of the matching b-cells are retained in circulation for a long time, acting as memory cells. The efficiency of the immune response to a given antigen is hence governed by the dynamically changing concentration of matching b-cells, which in turn depends on previous exposure to the antigen. In this way, the immune system adapts by building up high concentrations of b-cells that have proved useful in the past. Diversity is maintained by replacement of the cells in the bone marrow at the rate of about 5% per day, during which time mutation can occur.

### B. Idiotypic Network Theory

Jerne's idiotypic network theory [1] proposes that antibodies also possess a set of epitopes and so are capable of being recognized by other antibodies. Epitopes unique to an antibody type are termed idiotopes and the group of antibodies sharing the same idiotope belongs to the same idiotype. When an antibody's idiotope is recognized by the paratopes of other antibodies, it is suppressed and its concentration is reduced. However, when an antibody's paratope recognizes the idiotopes of other antibodies or the epitopes of antigens it is stimulated and its concentration increases. Jerne's theory hence views the immune system as a complex network of paratopes that recognize idiotopes and idiotopes that are recognized by paratopes, see Fig. 1. This implies that b-cells are not isolated, but are communicating with each other via collective dynamic network interactions [10]. The network is self-regulating and continually adapts itself, maintaining a steady state that reflects the global results of interacting with the environment [1]. This is in contrast to the clonal selection theory, which supports the view that change to immune memory is the result of antibody-antigen interactions only. In addition, Jerne's theory asserts that antibodies continue communicating even in the absence of antigens, which produces continual change of concentration levels. This can be interpreted as two forms of inter-antibody activity, "background" communication which occurs perpetually and "active" communication that takes place only when antigens are present. In the latter case, a single antibody becomes more dominant since the cell with the paratope that best fits the antigen epitope contributes more to the collective response [11]. It presents itself to the system as the *antigenic antibody* [7], which disturbs the network, inducing further inter-antibody suppression and stimulation.

*C. Incorporation of the Network Theory into Mobile Robotics*

Farmer *et al*. [2] propose that Jerne's hypothesis can be modeled as a differential equation simulating the changing concentrations of antibodies with respect to the stimulatory and suppressive effects and the natural death rate. Their model supposes that in a system with *N* antibodies [$x_1, x_2 ... x_N$] and *L* antigens [$y_1, y_2 ... y_L$], the differential equation governing rate of change in concentration *C* of antibody $x_i$ is given by (1).

$$\dot{C}(x_i) = b\left[\sum_{j=1}^{L} U_{ij}C(x_i)C(y_j) - k_1\sum_{m=1}^{N} V_{im}C(x_i)C(x_m) + \sum_{p=1}^{N} W_{ip}C(x_i)C(x_p)\right] - k_2 C(x_i) \quad (1)$$

The first sum in the square bracket expresses the stimulation of antibody $x_i$ in response to all antigens. Here, *U* represents a matching function between antibodies and antigens and the $C(x_i)C(y_j)$ terms model that the probability of a collision between them (and hence the probability of stimulation) is dependent on their relative concentrations. The second sum represents suppression of antibody $x_i$ in response to all other antibodies. *V* is a function that models the degree of recognition for suppression and $C(x_i)C(x_m)$ is the collision factor. The third sum models the stimulation of antibody $x_i$ in response to the other antibodies. The function *W* represents the degree of recognition for stimulation and $C(x_i)C(x_p)$ models the collisions. The variable $k_1$ allows possible inequalities between inter-antibody stimulation and suppression, but if $k_1 = 1$ these forces are equal. The $k_2$ term outside the brackets is a damping factor, which denotes the tendency of antibodies to die in the absence of interactions, with constant rate. Variable *b* is a rate constant that simulates both the number of collisions per unit time and the rate of antibody production when a collision occurs.

Equation (1) is based on the principle that antibody levels are dependent upon affinity between the antibody and the antigen, past use and the inter-antibody connections. The concentration levels are calculated dynamically in this way so that they can be used to determine fitness to the current environment. In addition, those with levels below a threshold can be eliminated from the system and replaced with new ones, as in nature.

Some robotics researchers construct communication networks without using the Farmer equation. For example Sathyanath *et al*. [23], [24] and Opp *et al*. [25] implement mine detection in the multi robot domain by modeling the locations of the mines and robots as the antigen epitopes and antibody paratopes respectively. A broadcast network that communicates antigen location information between the antibodies is analogous to the Jerne network. Robots are stimulated to move towards the mine to aid in diffusing it when they receive its location and are suppressed and move randomly otherwise. Idiotopes are not modeled and play no role in determining suppression and stimulation levels. In addition, the number of robots remains constant, meaning that variable antibody concentrations cannot be implemented.

However, most integrations of idiotypic selection and behavior-based robotics use the Farmer model since it approximates the biology very closely. For instance, Watanabe *et al*. [5], [6], [15] use the approach for a garbage-collecting robot with conflicting objectives. They represent epitopes, paratopes and idiotopes as binary strings that model the sensor readings, pre-condition and disallowed condition of the antibody respectively and use a roulette wheel manner of selection based on antibody concentrations after idiotypic interactions. The work presented in [15] is concerned with using reinforcement signals to derive appropriate idiotopes that are initially random. References [5] and [6] use a genetic algorithm with devised crossover to evolve the idiotopes, the network connections and the number and types of antibodies. Michelan and Von Zuben [12] solve the same problem, proposing a similar evolutionary mechanism for determining the network connections but they do not establish the antecedent and consequent parts of the antibodies automatically. Vargas *et al*. [7] also use the garbage example but evolve the network structure with a genetic algorithm and update the attributes that define their antibodies using a Learning Classifier System (LCS) [8]. Antibodies are selected based on activation, given by the product of concentration and strength of match to antigens after idiotypic effects have been calculated. Reinforcement learning is used both on the selected antibody and on those connected to it in the network. Krautmacher and Dilger [13] apply the idiotypic technique to navigation in a simulated maze world using the same basic approach as Watanabe *et al*. [5], but their antigens are variable, being composed of an object type and an object position and they do not implement meta-dynamics, i.e. antibodies are not replaced. Luh and Liu [9] use an idiotypic system to overcome local minima problems, modeling their antibodies as steering directions and their antigens as a fusion data set consisting of orientation of goal, distance between obstacles and sensors and positions of sensors. They implement stimulation and suppression by defining trigonometric relations between the steering angles.

In all the Farmer-based systems described ([5], [6], [7], [9], [12], [13] and [15]) antigens represent environmental situations, antibodies represent competence modules and the dynamics are governed using (1) or variations of it. However, the idiotypic controllers are not compared with base line systems to provide an indication of the idiotypic contribution to performance and no alternative selection procedures are tested. Furthermore, each paper assumes that the idiotypic system is readily adaptable to environmental change via highly flexible behavior selection, but the underlying mechanisms by which the dynamics facilitate selection of efficient and adaptable solutions are not explained in any depth.

## III. MOTIVATION

### A. Problems with the Reinforcement Approach

When robots explore terrain they are forced to make generalizations about environmental information and respond to those conclusions. For example, the message *object to right* could apply to a multitude of different situations, for instance, where another object is also fairly close to the left or a situation where turning away too much could lead the robot away from its target position. For this reason, a non-adaptive controller that prescribes a fixed course of action for each generalization will almost certainly lead the robot into a trap, i.e. into a position where it cannot free itself or repeats its behavior indefinitely.

Reinforcement learning, for example [16], [17] and [18] is more adaptive as it allows robots to score their performance and adjust their behavior accordingly, but it suffers from three main problems. First, the behaviors adopted and the speed of learning are too intimately linked with the reinforcement algorithms, which often need to be carefully engineered in order to yield a good solution. This compromises the system's autonomy. Second, the technique tends to undergo premature convergence preventing certain behaviors from being selected; a score increase is immediately awarded to the first successful behavior and other potentially better actions are hence perceived as inferior and subsequently ignored. Finally, when localized scoring structures are used, it can often take a long time for a robot to change its strategy when it gets caught in repeated behavior patterns that score positively in the local sense but do not contribute to achieving the overall goal. The delay is often caused by having to wait for an action's score to reduce sufficiently so that another is selected. If the reinforcement learning is not crafted carefully, robots can end up in never ending loops of repeated behavior.

## B. *The Idiotypic Advantage*

When Farmer-based idiotypic systems are implemented, behavior selection is a three stage process. The first stage is the nomination of the antibody with the highest strength of match to the presenting antigens (the antigenic antibody or stage 1 winner). In biological systems this degree of match is a physical attribute of the antibody's paratope, but in robotics where antibodies represent actions it is never accurately known and needs to be estimated, for example using current reinforcement learning scores.

During the second stage idiotypic suppression and stimulation occur. The antibodies with idiotopes that are recognized by the stage 1 winner's paratope are suppressed and those with paratopes that recognize the stage 1 winner's idiotope are stimulated. Earlier works have hinted that antibodies of the same type or species (i.e. valid "alternatives") should be chosen for stimulation and that different species should be suppressed. For example Watanabe *et al*. [5] suggest that stimulation and suppression chains work as a self and non-self recognizer. In addition Jerne [1] maintains that when an antibody paratope recognizes a foreign idiotope the suppressive forces dominate. This is not to say that antibodies identical to the stage 1 winner should be stimulated and others suppressed because this would exacerbate premature convergence. The main function of idiotypic communication is to promote those antibodies that demonstrate a balance between similarity with and difference from the first winner. Simplistically this can be viewed as stimulation of antibodies of the same basic type (or species) but possessing different parameters. For example, reversing backwards in a straight line and reversing with a left spin of 30º are both of type "reverse" but have different spin components. Stimulation increases strength of match and suppression reduces it so that the antibody with the highest strength of match after these effects have been calculated has a high chance of being selected to execute its action. The actual antibody chosen also depends on the third stage, which considers current antibody concentration levels as well as the strength of match. In some cases the final elected antibody is the stage 1 winner; in others a different antibody may be called.

Theoretically, one should see improvement in a robot's performance when idiotypic suppression and stimulation are introduced into reinforcement learning based behavior selection. This is because the idiotypic system is potentially able to overcome the three main problems listed above. Although the system is still highly dependent on the structure of the reinforcement learning (since antibody-antigen matrices are updated according to the reinforcement scores awarded), the action with the highest stage 1 fitness is not always selected and the concentrations of all antibodies are adjusted according to the degrees of stimulation and suppression. This should instigate a degree of detachment from the engineered learning, providing a more autonomous approach. In addition, the method should significantly reduce premature convergence since antibodies with lower stage 1 fitness should also get a chance to demonstrate their abilities and increase their fitness. This offers increased flexibility to derive more creative solutions to problems. In addition, robots should be able to break out of repeated sets of behavior much faster since they do not have to wait for fitness to reduce before another behavior is selected. The idiotypic network should provide a more dynamic system that demonstrates a higher rate of antibody change, potentially enabling the robot to break the cycle. Even if the cycle is not broken straight away, the dynamics should ensure that a suitable behavior is eventually chosen.

*C. Hypotheses on the Performance of Idiotypic Networks*

As stated above, an idiotypic network should be able to overcome the problems associated with reinforcement learning. To this end, three hypotheses are proposed as follows:

$H_1$  The idiotypic network system shows a degree of de-coupling from engineered reinforcement learning and hence provides a more autonomous approach.

$H_2$  The idiotypic network system significantly reduces the problem of premature convergence.

$H_3$  The idiotypic network system allows rapid escape from repeated behavior patterns or prevents them from happening entirely.

Note that $H_3$ is linked to $H_2$ since reduced premature convergence facilitates a less greedy strategy, which encourages more varied behaviors. The following sections describe the problem, models, programs and experimental procedures that are used to test the above hypotheses.

## IV. TEST ENVIRONMENT AND PROBLEM

The agents used in this research are virtual Pioneer P3DX robots that are required to navigate around maze worlds developed with Stage 2.0.1, a 2D simulator for the Player 2.0.1 interface [19]. For example, *Maze World* represents a fictitious building in which the robot must travel through six rooms A – F, avoiding obstacles and entrapment (see Fig. 2). Small square cyan markers are used to indicate the doorways and competence modules for detecting them with a camera and tracking them are provided. Once the robot has passed a marker or doorway, the path back is manually blocked using the movable blocking lines shown in Fig. 2. The blocking positions are also indicated using dashed lines. This procedure effectively simulates automatic closure of the doors once the robot has passed through.

The robot carries a SICK LMS 200 laser with minimum range set at 0.0 m and maximum range set at 8.0 m. The device takes 361 readings covering the front 180° and measures the distances between the robot's centre and any obstacles ahead. When processing the data this area is divided into eight equal sectors 0 – 7 each 22.5° wide, with sectors 3 and 4 at the front of the robot, sectors 0, 1 and 2 to the left and sectors 5, 6 and 7 to the right. The minimum reading and its sector, the sector with the maximum reading and the average reading are determined. A Canon VCC4 pan-tilt-zoom camera and blob finding software are used to search for the door markers. The blob finding software enables translation of the camera data into groups of like-colored pixels or blobs distinguishable by their RGB value. The camera remains fixed ahead at 0° at all times, with field of view set to 60°. The internal odometry determines the distance traveled and eight rear sonar measure the average distance behind the robot.

The robot is started in room A1 in the position shown, with its final target mid-way through room F, i.e. it is allowed to stop when the blob area from the final line is greater than 1,000 pixels. The robot's performance is assessed according to how fast it completes the journey and by the number of collisions with the obstacles or walls. Additionally, *Mirror World*, a mirror image of *Maze World* is used to test the robot's performance after initial training has been carried out.

The mazes are deliberately designed to facilitate the drawing of more general conclusions, i.e. the problems are entirely solvable but provide a level of difficulty suitable for differentiating between weak and strong methodologies. For example, the doors are wide enough for the robot to pass through, but small enough so that very refined movements are required if the robot is to pass without collision. Obstacles are strategically placed so that doorways are not blocked, but also so that freedom of movement is restricted in some places. The course is kept fixed throughout all experiments to provide a fair comparison between the different approaches. Although it may be argued that variation of the environment is limited, there are several rooms in the world and each of these may be considered a sub-environment. Furthermore, the worlds used have proved extremely non-deterministic.

The control software uses the libplayerc++ client library developed for use with the Player server (version 2.0.1) and it is run on GNU/Linux 2.6.9 (CentOS distribution) with a Pentium 4 processor and 3.6 GHz of memory. All simulations are run in real time.

## V. SYSTEM ARCHITECTURE

### A. *Equations Used to Model the Network*

The functions *V* and *W* in (1) model both background antibody communication and active antibody communication, i.e. they compare each antibody with all of the others so that the levels of stimulation and suppression can be determined. Background communication is not simulated here as active communication represents a stronger force and since in this system all environmental situations are modeled as antigens (even the case where average sensor readings are high), active communication is of most interest. In addition the removal of background communication produces a simpler system, as each antibody need only be compared with the antigenic antibody denoted here as $x_{w1}$. The communication is mimicked by comparing the paratope of $x_{w1}$ with the idiotopes of the other competing antibodies and vice versa. This involves constructing a paratope matrix $P$ that shows the strength of match between antibodies and antigens and an idiotope matrix $I$ that shows disallowed matches so that desired combinations can be recognized against unwanted ones. The process of computing the internal network effects thus consists of summing $P$ and $I$ strength of match values between antigens and antibodies. A variation of Farmer's equation (2) that sums the inter-antibody suppressive and stimulatory effects over the number of antigens $L$ rather than the number of antibodies is hence used. The idiotypic matching functions are termed $V'$ and $W'$ here to distinguish them from functions $V$ and $W$ in (1). Another important difference is that equation (2) uses concentration of the antigenic antibody $C(x_{w1})$ rather than that of every antibody in the system $C(x_m)$ and $C(x_p)$ in (1). In addition, the antigen concentration term $C(y_j)$ and the associated $C(x_i)$ term in (1) have been removed from the first part of (2) since antigen concentrations are not modeled and their relative importance is already represented by weighting them according to a priority ranking. Other than this, the terms in (2) are identical to those used in (1), which are fully explained in Section II, Part C.

$$\dot{C}(x_i) = b\left[\sum_{j=1}^{L} U_{ij} - k_1 \sum_{m=1}^{L} V'_{im} C(x_i) C(x_{w1}) + \sum_{p=1}^{L} W'_{ip} C(x_i) C(x_{w1})\right] - k_2 C(x_i) \quad (2)$$

Equation (2) must be evaluated in separate parts, since the antigenic antibody $x_{w1}$ is unknown until the first sum in the square brackets is used to determine it. It is therefore split into five separate equations (3) – (7). Equation (3) represents the first sum in the square brackets, i.e. $T_1(x_i)$ is the strength of match of antibody $x_i$ to the set of presenting antigens and $U$ is the antigen matching function. Once (3) is evaluated for all antibodies, the antibody with the highest $T_1(x_i)$ value is selected as the antigenic antibody $x_{w1}$, therefore equation (3) is always processed first. Equation (4) represents the second sum in the square brackets, i.e. $T_2(x_i)$ calculates the suppression in antibody $x_i$ by using $V'$ as a suppressive matching function and modeling the probability of collisions between antibodies $x_i$ and $x_{w1}$. Similarly, $T_3(x_i)$ in (5) sums the stimulation in $x_i$, using $W'$ as a stimulatory matching function. Functions $U$, $V'$ and $W'$ are expressed in terms of $P$ and $I$ and are explained further in Part D. $T_g(x_i)$ in (6) represents the global strength of match, a strength metric that encompasses all molecular activity, i.e. $T_g(x_i)$ equates to all of the terms in the square bracket in (2). The parameter $k_1$ in (6) is the same as in (1) and (2).

Equation (7) is equation (2) re-written in terms of $T_g(x_i)$ and it expresses the rate of change of concentration of antibody $x_i$ with time. As the concentrations must be computed discretely, a difference equation form of (7) is used, (8).

$$T_1(x_i) = \sum_{j=1}^{L} U_{ij} \tag{3}$$

$$T_2(x_i) = \sum_{m=1}^{L} V'_{im} C(x_i) C(x_{w1}) \tag{4}$$

$$T_3(x_i) = \sum_{p=1}^{L} W'_{ip} C(x_i) C(x_{w1}) \tag{5}$$

$$T_g(x_i) = T_1(x_i) - k_1 \big(T_2(x_i)\big) + T_3(x_i) \tag{6}$$

$$\dot{C}(x_i) = b\big(T_g(x_i)\big) - k_2 C(x_i) \tag{7}$$

$$C(x_i)_{t+1} = C(x_i)_t + b\big(T_g(x_i)_t\big) - k_2 C(x_i)_t \tag{8}$$

In hybrid AIS systems antibody fitness $F$ is often measured using a combination of metrics that represent the individual components of the scheme. For example [7] uses activation $A(x_i)$, defined as the product of the global strength of match $T_g(x_i)$ and the concentration, see (9). This method of assessing fitness is adopted here as it incorporates both the AIS and reinforcement learning aspects of the hybrid system.

$$A(x_i) = C(x_i)_{t+1} T_g(x_i) \tag{9}$$

### B. Choice of Antibodies and Antigens

As in many previous robot AIS systems ([5], [6], [7], [9], [12], [13] and [15]), environmental situations are modeled as antigens and competence modules are modeled as antibodies. For simplicity, fixed numbers are used (8 antigens and 16 antibodies) and antibody replacement is not implemented.

The set of antigens (listed in Table I) is given a priority structure based on the principle that the needs of some situations outweigh those of others. For example, if the robot is stalled against a wall it must take action to free itself before it can deal with less urgent problems such as an object to the left. Since antigens 0 – 6 have a wide application to most robot navigation problems and antigen 7 would be useful for any problem involving tracking an object, this priority ranking is reasonably unspecific and means that more general conclusions may be drawn from the experiments. The condition parameters are selected from the results of conducting pre-trials that enabled the door tracking task to be carried out efficiently using system S1.

The antibody repertoire, i.e. the set of possible behaviors listed in Table II, is selected on the basis of providing the robot with the ability to move in a number of different directions and at a number of different speeds covering both the front and rear. In addition, selection of more intelligent actions such as wandering toward the maximum laser reading and tracking the door markers are provided. Apart from tracking the markers, all the actions may be considered generic to navigation problems that require a robot to avoid obstacles and traps. The maximum speed permitted $M$ is 2.0 m/s.

*C. The Paratope and Idiotope Matrices*

Five different paratope matrices $P_1$ - $P_5$ (antibody-antigen strength of match matrices) are used in the experiments. To help minimize any initial bias these are prepared beforehand by generating random element values $P[x_i, y_j]$ between 0.50 and 0.75, i.e. not too high and not too low. These values are then adjusted by adding a small number $\delta(x_i)$ (positive or negative) to each antibody's elements so that the mean across each row of $P$ is 0.625. Variable $\delta(x_i)$ is given by (10) where the original mean for row $i$ is denoted by $\mu_i$ and the desired mean by $\mu_d$. Again, $L$ is the number of antigens. The derivation is given below.

$$L\mu_d = \sum_{j=1}^{L}\left(P[x_i, y_j] + \delta(x_i)\right) = \sum_{j=1}^{L}\left(P[x_i, y_j]\right) + L\delta(x_i)$$

$$L\mu_i = \sum_{j=1}^{L} P[x_i, y_j] \qquad L\mu_d = L\mu_i + L\delta(x_i)$$

$$\delta(x_i) = \mu_d - \mu_i \tag{10}$$

When the program executes the elements of $P$ are updated approximately once each second using reinforcement learning. However, it is worth noting that values are not allowed to fall below 0.00 or rise above 1.00.

Only one fixed idiotope matrix $I$ is used, i.e. it is not permitted to change, either within the duration of the robot's run or throughout the course of the experiments. This is deliberate in order to render easy investigation and explanation of the idiotypic mechanisms. The matrix is hand coded according to perceived disallowed antibody-antigen combinations, i.e. pairs that would produce non-sensible or unwanted actions are given positive element values, see Table IV. Numbers in the range 0.00 to 1.00 are possible, but the sum of elements for each antibody (across all antigens) is set to 1.00. This is to reduce the likelihood of any antibody becoming over stimulated or suppressed. The positive values indicate the level of confidence that the combination is poor in some way. Null values do not necessarily indicate a good combination, they merely show complete uncertainty.

*D. The Algorithm and Matching Functions*

The random, variable paratope matrix *P* and fixed idiotope matrix *I* are both imported from files and the robot sensors are read in a continuous loop. The system checks the sensor data for the presence of antigens approximately once per second, i.e. every ten iterations of the continuous loop.

Multiple antigens are allowed to present themselves simultaneously, but one is deemed dominant according to the priorities given in Table I. An antigen array $G(x_i)$ is formed, which has value 0 for non-presenting antigens, value 2 for a dominant antigen $y_d$ with $P[x_i, y_d] > 0$, value 0 for a dominant antigen with $P[x_i, y_d] = 0$ and value ¼ for all other presenting antigens, so that the dominant one receives greater weighting for all antibodies with positive $P[x_i, y_d]$. For example, if antigens 2, 4 and 7 present themselves then $G(x_i) = [0, 0, ¼, 0, 2, 0, 0, 0, ¼]$, provided $P[x_i, y_d] > 0$.

An antibody is selected to have its action executed in response to the presenting antigens and this is effectively a three stage process. The first stage is selection of the antigenic antibody $x_{w1}$ by computing $T_1(x_i)$, i.e. summing strength of match to the antigen set using (3), where the matching function *U* is defined by (11) below. Here *P* is the paratope matrix and *G* is the antigen array.

$$U_{ij} = P[x_i, y_j]G(x_i)_j \qquad (11)$$

This definition of *U* uses the weighted current reinforcement scores for antigen matching and ensures that all antibodies with zero match to the dominant antigen are discounted, i.e. are assigned $T_1 = 0$. This is important since reinforcement learning operates on the degree of match to the dominant antigen. Negative scoring would therefore have no effect on antibodies with zero match since match values are not allowed to fall below zero. Once $x_{w1}$ is determined it presents itself to the idiotypic network as the antigenic antibody.

The second stage is summing the stimulation and suppression that the antigenic antibody causes. In other words each antibody's global strength value $T_g(x_i)$ is calculated by computing the $T_2(x_i)$ and $T_3(x_i)$ values from (4) and (5) that represent the effects of suppression and stimulation respectively. In this algorithm all other antibodies with $T_1 > 0$ must compete with $x_{w1}$ for selection via the idiotypic process. Note that $x_{w1}$ does not compete with itself, i.e. it is not permitted to stimulate or suppress itself; its strength remains unchanged throughout the entire second stage, i.e. $T_g(x_{w1}) = T_1(x_{w1})$, $T_2(x_{w1})$ and $T_3(x_{w1}) = 0$. In addition, non-competing antibodies must have $T_g = T_1 = T_2 = T_3 = 0$. For this purpose, an antibody array $H$ is formed that has value 1 for competing antibodies with $T_1 > 0$, but value 0 for antibodies with $T_1 = 0$ and antibody $x_{w1}$. The function $V'$ in (4) is given by (12) below, where $I$ is the fixed idiotope matrix.

$$V'_{im} = P[x_{w1}, y_m] I[x_i, y_m] H_i \tag{12}$$

Under this definition of $V'$ equation (4) simulates suppression by comparing the stage 1 winner's paratope with the competing antibody's idiotope. Since the paratope constitutes approved antigen matches and the idiotope shows disallowance, the product of these elements provides a good indication of the level of suppression that should be induced in the competitor.

$$W'_{ip} = (1 - P[x_i, y_p]) I[x_{w1}, y_p] H_i \tag{13}$$

The function $W'$ in (5) is given by (13). This definition allows equation (5) to mimic stimulation, this time examining the stage 1 winner's idiotope and the competing antibody's paratope. A low paratope element coupled with a positive idiotope value indicates a possible similar species and that the antibody should be stimulated. Here, the paratope element is subtracted from 1 in order that the elemental product yields high values for a high level of stimulation.

Equations (4) and (5) show that the elemental products in (12) and (13) are summed over all the antigens and multiplied by the concentration terms. In this way an individual antibody may undergo multiple idiotypic suppressions and stimulations. The net result of these and the original antigen matching determines $T_g(x_i)$ via (6).

$$\|C(x_i)\|_{t+1} = \frac{C(x_i)_{t+1}}{\sum_{j=1}^{N} C(x_j)_{t+1}} \tag{14}$$

The third stage involves the use of (8) to calculate each new antibody concentration and (9) to calculate activation. Here, the term concentration is used to mean the proportional number of clones of an antibody type in circulation if the total number is held at $N$, where $N$ is the number of antibodies. Therefore all 16 antibodies begin with equal concentrations of 1. Once the new concentration values are derived from (8) they are normalized using (14), and multiplied by $N$ to keep the total number of clones at 16. This process prevents scaling problems that arise when the total number becomes too high and is in keeping with many other investigations involving AIS networks, for example [6] and [21]. In addition, studies with mice have suggested that an almost constant number of b-cells are active, so it is likely that there is a mechanism in nature that controls this [20].

In order to test the hypotheses $H_1$ - $H_3$, an incremental approach is adopted, i.e. three experimental systems S1 – S3, each with increasing levels of idiotypic complexity are used to solve the navigation problem and the performance of each is compared in terms of the speed and agility of the robot. S1's overall winning antibody is always $x_{w1}$ but S2's is the antibody with the highest global strength of match $T_g(x_i)$ and S3's is the one with the highest activation $A(x_i)$. S1 and S2 are therefore simply sub-programs of S3 that implement only stages 1 and 2 respectively. In addition, S1 and S2's antibody concentrations are held constant at 1 throughout, i.e. the terms $C(x_i)$ and $C(x_{w1})$ in (4) and (5) are effectively ignored to simplify the dynamics and provide an indication of the effect of introducing a network that is independent of concentration. However, in S3 the concentration levels are allowed to vary, i.e. the system implements the concentration terms in (4), (5), (8) and (9) as variables, which represents a full AIS system. Table III summarizes the three systems and describes how fitness is measured for each.

Since fitness in S1 does not consider idiotypic effects and ignores concentrations of antibodies it is purely an RL system that uses $P$ as a belief table for executing actions. It is in no sense an AIS. S2 is not a true AIS as it does not base selection on a function of antibody concentration and molecular collisions are not modeled within the network. In addition, the system has no global feedback from the network as the strength $T_g(x_i)$ is used only to select the fittest antibody, and only the fittest undergoes a paratope adjustment from reinforcement learning. The $T_g(x_i)$ values for the other antibodies are not used to adjust the paratope in any way. System S3 represents a true AIS because feedback from the network is global through alteration of all antibody concentrations using (8) and there is also feedback from concentrations to the network since collisions between molecules are modeled in (4) and (5).

To illustrate use of the equations (3) – (6), Table VI shows the results of calculating $T_1(x_i)$, $T_2(x_i)$, $T_3(x_i)$ and $T_g(x_i)$ using the idiotope values from Table IV and the example paratope values given in Table V. In these calculations all antibody concentrations are held constant at 1 for simplicity, as in the case of S2 and $k_1$ is set at 0.625. In the example, the antigens presented to the system are 1, 3 and 5, hence 5 is dominant. The table shows that the stage 1 winner is antibody 14 but that the idiotypic processes nominate antibody 10, i.e. an alternative reverse antibody as the overall winner.

Once the fittest antibody has been chosen it executes its action and its effect is assessed using reinforcement learning. The appropriate element of the paratope matrix $P$ is either increased or decreased, depending on whether a reward or penalty is issued by the reinforcement algorithm.

*E. Reinforcement Learning*

Reinforcement learning occurs when knowledge is implicitly coded in a scalar reward or penalty function. There is no teacher and no instruction about the correct action, just a score that is yielded by the robot's interaction with its environment. Here, the technique is employed for dynamic estimation of the degree of match between antibodies (actions) and antigens (environmental situations).

This paper seeks to compare a basic RL system (S1) with hybrids that utilize an idiotypic network (S2 and S3) and thus establish whether the idiotypic treatment enhances robot performance. It is therefore essential to construct a good reinforcement scheme to test whether the network can add something to a design that already performs reasonably well. For this reason, a scheme that is highly engineered is used. This is based on a "brute force" approach that coerces the robot into behaving in a desirable way, e.g. by penalizing it for going backwards under certain conditions. The reward and penalty increments coded are ratio orientated e.g. the robot is rewarded more when it travels fast than when it travels slow if the sensors show no danger.

As well as testing a good RL algorithm, a weaker strategy is also employed as a direst test for $H_1$. If idiotypic robots can produce good results despite using poor learning then they will have demonstrated a degree of detachment from the structured reinforcement signals.

In both strategies the reinforcement value $r_f$ is set to 0 at the start of the learning exercise, which is carried out once every ten loops but five loops out of synchronization with the completion of the actions. In other words, approximately half a second after acting, the selected antibody's performance is scored either negatively or positively by re-reading the sensor values and using one of the scoring algorithms described below. The algorithm used is largely dependent upon the dominant antigen. However, this does not render the scheme too problem specific because the antigens represent environmental situations that are reasonably universal to navigation and tracking problems. A brief description of the stronger reinforcement design is given below. Note that the absolute values of the scoring parameters are not presented as they are somewhat arbitrary. The network system S3 should be able to work alongside any basic RL scheme to enhance performance.

For dominant antigens 0, 1 and 2 (*obstacles present*) the learning algorithm provides linear rewards for an increase in the distance between the robot and the obstacle and linear penalties for a decrease. However, if the sector producing the minimum reading changes it is not immediately obvious whether the robot is encountering the same object, so the scores are reduced by a factor of 4. Reverse antibodies are awarded an additional penalty since reversing away from obstacles does not contribute to the overall goal of moving forwards towards the target.

For dominant antigen 4 (*low average laser reading*) the algorithm scores in a linear fashion, rewarding an increase in average laser reading and penalizing a decrease. As one of the objectives of robot navigation is to utilize space so that collisions are avoided it also makes sense to reward any antibody that is able to move the robot forward from enclosed to more open areas. The change in average reading is hence also used as a global assessment metric for all cases, regardless of the dominant antigen. In addition, reverse antibodies that have reduced the average reading are penalized further to discourage their general use. Reversing is contrary to the overall goal and should only be necessary to escape from stall situations.

When antigens 5 or 6 (*stalled* or *blocked behind*) are dominant assessment is based on the distance traveled in the half second between reading the sensors and scoring. This scheme provides linear rewards for movement and a fixed penalty for failing to move.

The algorithm for dominant antigens 3 and 7 (*average reading above threshold* and *door marker seen*) rewards faster antibodies as the robot can afford to travel quickly when no obstacles are present and it is not trapped. Slower antibodies receive a small penalty. Additionally, antibodies that keep the door marker in sight are rewarded further and the score is even greater for those moving directly toward it. In contrast, antibodies close to a door marker that then lose sight of it are penalized. Again, $r_f$ is reduced for negative speeds.

The final reinforcement learning score $r_f$ (either positive or negative) is added to the paratope matrix element $P[x_w, y_d]$, i.e. that representing the affinity between the dominant antigen $y_d$ and the overall winning antibody $x_w$. However, if $P[x_w, y_d]$ becomes negative as a result of this, it is set to 0. The algorithm is summarized by (15) below.

$$P[x_w, y_d]_{t+1} = \text{MAX}\left(0, P[x_w, y_d]_t + (r_f)_{t+0.5}\right) \tag{15}$$

Any antibodies that are penalized also have their concentration increase removed, i.e. their concentration is set back to the figures from the previous iteration.

The weaker learning strategy is the same as that described above except that it over penalizes the obstacle avoidance antibodies 1, 2, 4, 5, 6, 7, 8, 9 and 12 by applying the door tracking part of the algorithm for dominant antigens 3 and 7 to all antigens. This is too tough a test and as a result robots do not tend to develop very good obstacle avoidance strategies. The program architecture also differs slightly as all dominant antigens, even those with zero $P[x_i, y_d]$, are given a value of 2 in the array $G(x_i)$ in (11). The system is thus less robust since antibodies with zero match to the dominant antigen may be selected, meaning that negative reinforcement scoring has no effect. This encourages repeated behavior.

## VI. EXPERIMENTAL PROCEDURES

### A. Measuring Robot Performance and System Properties

The program recognizes a collision when the dominant antigen is either *blocked behind* or *stalled*. However, on its own the number of collisions or stalls $n_s$ does not represent a good measure of task fitness since it does not allow for robots that are highly cautious but too slow. Robots should be able to complete the course as rapidly and with as few stalls as possible. On the other hand, the time to complete the course $t$ does not provide an indication of the robot's safety attributes; a fast robot is no good if it damages itself or the environment. For this reason a score metric $S$ that combines the run time with the number of stalls is used to determine task fitness. $S$ is defined by (16), where $\varphi$ is the ratio of mean $t$ to mean $n_s$ over all experiments (17). The score thus gives equal weighting to $n_s$ and provides a linear combination of the two metrics that has the same mean as $t$ over all runs, see proof below. Throughout these experiments $\varphi$ is set at 9.08, the figure computed in a series of pre-trials using all three systems, S1 – S3.

$$S = \frac{\varphi n_s + t}{2} \tag{16}$$

$$\varphi = \frac{\bar{t}}{\bar{n}_s} \tag{17}$$

$$\bar{S} = \frac{\varphi \Sigma n_s + \Sigma t}{2N} = \frac{\varphi \bar{n}_s + \bar{t}}{2} = \frac{\bar{t} \bar{n}_s}{2\bar{n}_s} + \frac{\bar{t}}{2} = \bar{t}$$

In addition to *S*, the reinforcement success rate, defined as the percentage of positive scores awarded and the rate of antibody change (the percentage of selected antibodies that differ from the previous iteration) are measured for each system. For S2 and S3 the rate of idiotypic difference (the percentage of iterations where the stage 1 winner and final winner differ) and reinforcement success rate when an idiotypic difference occurs are calculated, both over the entire run and during stall conditions only. These metrics may help in explaining any perceived differences between the systems. A set of data that shows the iteration number and antibody used during a collision is also preserved.

*B. Selection of Parameters – Initial Investigation*

Before exhaustive comparisons of the three systems are carried out, several preliminary investigations are undertaken to establish suitable parameters for *b* and $k_1$ in equations (6) and (8). However, since all antibodies are to be retained in the repertoire without replacement, it is not necessary to test so stringently for an acceptable value for $k_2$. This parameter serves mainly to determine how rapidly antibodies die out in system S3, so as long as it is kept low in comparison to *b* no antibodies are removed from the system. Testing has shown that a value of 0.05 is effective for this purpose when the system is implemented with $8 \leq b \leq 800$ and with $0.00 \leq k_1 \leq 1.00$. If a meta-dynamic system were to be used, $k_2$ would become much more important, see [21]. As antibody replacement is not the focus of this work, the value of 0.05 is retained throughout all further experiments.

In order to establish satisfactory values for *b* and $k_1$ the *Maze World* environment is used as the test bed with the robot using the initial paratope matrix $P_1$. As a precursor to a more thorough treatment, first $b$ is set at 8 for system S3 and a set of results for $k_1$ values between 0.0 and 1.0 are determined by finding the mean score $S$ for six runs. Next $b$ values of 80 and 800 are trialed in the same way and a set of mean scores for system S2 and a mean score for system S1 are found. For S1 30 runs are completed and a 95% confidence interval is computed using a standard one-tailed t-test. The mean rate of idiotypic difference is also measured for systems S2 and S3 each time. Fig. 3 shows a plot of mean score against $k_1$ for S2 and for S3 using the different $b$ values and Fig. 4 shows mean idiotypic difference rate. Although system S1 does not use $k_1$ its performance is included in Fig 3. as a straight line for comparison purposes.

The chart in Fig. 4 shows that $k_1$ has a great effect on the degree of idiotypic influence for both S3 and S2 and strongly suggests that this is independent of the value used for $b$ in S3. It also shows that $k_1$ values in the range 0.0 to 1.0 produce difference rates almost between 0% – 90%, although after $k_1 = 0.8$ this tails off to about 4%. In fact, the relationship between $k_1$ and idiotypic difference rate appears almost linear, which fits in with the theory since from (6) a low $k_1$ should reduce the suppression of antibodies, producing a high number of iterations where the antibody selected first differs from the final winner. In contrast, a higher value should increase suppression and reduce the rate of difference. It is also notable that for $k_1$ values lower than about 0.8 S2's rates are slightly lower than those of S3.

The chart in Fig. 3 provides clear evidence that the performance of the robot is dependent upon selecting appropriate values for both $k_1$ and $b$. The 95% confidence interval for the mean score of S1 is between 402 and 265, which means that out of the four other systems trialed, only S3 with $b$ set at 8 or 80 is likely to be able to produce results with significantly better scores. System S2 and system S3 with $b = 800$ both stay within these confidence limits for all values of $k_1$ tested. In addition, for $b = 80$ and $b = 8$ the region of the x-axis between 0.45 and 0.65 shows a trend towards a dip in mean score, which remains well below the lower limit for S1.

The evidence from Fig. 3 and Fig. 4 suggests that there is an optimal level for the idiotypic difference rate and hence for $k_1$; a low level of suppression (low $k_1$) produces a system that tries alternative behaviors too often as it is not strict enough in its selection of them, but a high suppression level (high $k_1$) creates a system that does not try them often enough as it is too rigid. Hence a system with a higher $k_1$ value has a lower idiotypic difference rate and as the $k_1$ value becomes higher it becomes less distinct from system S1, which has 0% difference since it does not use idiotypic selection. This theory is supported by Fig. 3 as the lower $k_1$ values remain within the bounds for the mean score of S1 for all the systems tested. Alternatively, as $k_1$ approaches zero the robot will tend not to accept the winner from stage 1 and will hence miss out on striking a desirable balance between accepting and rejecting it. Consequently, its performance deteriorates as evidenced in the graph.

Fig. 4 shows a remarkably similar pattern of idiotypic difference for widely different $b$ values, suggesting that $b$ does not have much influence on idiotypic stimulation and suppression levels. However it is clear from Fig. 3 that a $b$ value of 800 is not likely to out-perform S1 for any given $k_1$ value, so how can this be explained? From (8) it is apparent that $b$ plays an important role in determining the weighting that is given to the global strength of match $T_g$ when calculating the new concentration values. It is therefore used to provide an indication of the relative importance of $T_g$ against the historic concentration value $C(x_i)_t$. Use of lower values favors antibodies that have been successful in the past, whereas a high value tends to produce a system that chooses those that best match current environmental information. In addition, a higher value gives rise to a faster rate of change of concentration meaning that levels can build up and reduce rapidly between iterations, providing a more useful indication of antibody fitness. It is likely that there is a range of values that strike a good balance between using historical data and current environmental information and it is probable that 800 is too high for this range when implementing this reinforcement structure with this particular environment and idiotope.

There is an extremely high level of correlation between the score data, stall data and task completion time data, i.e. the patterns in the graphs are almost identical. For this reason, it is sufficient to proceed by examining score data only.

## C. Selection of Parameters – More Detailed Investigation

In order to gain a better understanding of the role of $b$ and $k_1$, a wider range of $b$ values (8, 15, 20, 40, 60, 80, 100, 120, 160, 200 and 800) is examined. This range is used for $k_1$ between 0.45 and 0.65 in increments of 0.50, adopting the same experimental procedure, i.e. six runs in *Maze World* starting with the random paratope matrix $P_1$ only. This region of $k_1$ is selected because of its superior performance in the first set of trials, i.e. it is assumed that $k_1$ values outside of this range are unlikely to yield mean scores significantly different to those from S1.

Fig. 5 shows mean scores against $b$ for $k_1$ values in the selected range. It is readily apparent from the graph that there is a region of $b$ where best performance is with $k_1$ set at 0.60 and where best performance also represents a "good" performance with mean score less than 200. This range is approximately between 40 and 160. When $k_1 = 0.60$ is used with higher values of $b$, performance drops considerably with no best scores lower than 200. At the other end of the scale, lower $b$ values still perform well, but the optimum $k_1$ appears to have shifted towards a lower value.

Fig. 6 is a plot of mean score against $k_1$ for the region $40 \leq b \leq 160$ only. Here, maximum performance occurs at $k_1 = 0.60$ for each system and mean score values are well below the lower confidence limit line for S1. The same is not true for any of the other values tested, although it is possible that values very close to 0.60 may also produce this phenomenon. In addition, one-tailed t-tests have shown that the mean score (173) at $k_1 = 0.60$ in this range of $b$ is significantly higher than the mean scores for all the other values of $k_1$, (295, 254, 248, 234) at the 99% level. It is reasonable to assume that the optimum value for $k_1$ is very close to 0.60 for this scheme, environment and idiotope, as long as $b$ remains in the approximate range 40 to 160.

An interpretation of the graphs in Figs. 5 and 6 is that the approximate region $40 \leq b \leq 160$ represents a more stable form of equation (8), which shows optimum performance close to $k_1 = 0.60$, i.e. when the idiotypic difference rate is at about 20%. The stability is attributed to the selection of $b$, which permits a good balance between use of historical antibody information and reaction to the environment. However, for higher $b$ values, the equation becomes too dependent on environmental information, i.e. it becomes more like S2 since concentration plays a less important role. It is not surprising that the performance of S2 and S3 with $b = 800$ are similar in Fig. 3. Conversely, for $b$ less than 40 the system tends to rely more on historical information, placing less emphasis on which antibody has the higher $T_g$ value and more on concentration.

These preliminary experiments provide a good gauge for parameter setting during more extensive testing of S3 and have also shown that $b$ is far more robust than $k_1$ since good performances can be achieved for widely differing $b$ values, whereas $k_1$ is much more sensitive to change. In addition, pre-trials with the weak learning strategy have revealed that the rate of idiotypic difference is intrinsically higher than the strong learning strategy because the robot is penalized by reinforcement more often. Consequently the "optimum" $k_1$ value increases for the stable region of $b$ when learning is weaker. This is because more suppression is needed to yield an equivalent idiotypic difference rate to the stronger learning strategy.

*D. More Rigorous Comparison of S1, S2 and S3*

Here, the aim is to show that the AIS-RF hybrid S3 can perform significantly better than the RF system S1 or indeed the simpler hybrid S2. When investigating parameters six runs are able to provide a strong indication of robot performance, but six runs are not sufficient for accurately testing significant differences between the systems. For this reason, each system S1 – S3 is tested 30 times in *Maze World*, six times with each of the five initially random paratope matrices $P_1 – P_5$. The 90 runs are repeated again in *Mirror World*, starting not with a random paratope matrix, but with the appropriate matrix saved from the first run. This is to test how well the robot learns from its first experience and to distinguish a "lucky" first run from a genuinely resourceful one, where a good set of behaviors develops.

The parameter $k_1$ is set at 0.625 for both S2 and S3 as this represents a value close to the empirically measured "optimum". In addition, the parameter $b$ is set at 80 for S3 as this value lies well within the stable region.

Results measure mean, maximum and minimum number of stalls, task completion time and score and also mean antibody change rate, idiotypic difference rate and reinforcement success rate. Runs with scores below 200 are labeled as good since they show above average attainment. Alternatively, runs with scores greater than 400 are declared bad since this represents a performance ranked in the bottom 10%.

However, the above measurements provide information on all runs and do not indicate what is happening to system dynamics when robots get into difficulty, i.e. when they are spending a lot of time trying to escape entrapment. For this reason, a number of runs with an $n_s$ value higher than average, (i.e. a lower performance) are sampled from the S1 and S3 data, so that antibody information is available for approximately 80 long stall sequences, i.e. sequences that last more than one iteration.

It is also important to establish that S3 is able to out-perform S1 for other values of $k_1$ and $b$ within the established "optimum" regions. For this reason an additional two sets of 30 runs are conducted with S3 in *Maze World* using untrained robots. These two tests use $k_1 = 0.600$, $b = 60$ and $k_1 = 0.585$, $b = 100$. S2 is not tested further with different parameters since it does not use $b$ and its preliminary results already covered the region $0.0 \leq k_1 \leq 1.0$. In addition, S1 and S3 are the systems of most interest.

Finally, S1 and S3 are tested with untrained robots in *Maze World* using the weak reinforcement learning strategy. Only one world is used since only one comparison with the good scheme is necessary. Although the original value used for $b$ (80) is preserved, $k_1$ is raised to 0.800 to reduce the idiotypic difference rate down to a level comparable to the rate for the "optimum" $k_1$ value used in strong learning experiments, i.e. to about 20%.

## VII. RESULTS AND DISCUSSION

Table VII shows the means and standard deviations for $S$, $t$ and $n_s$ and the percentages of good and bad runs for each system used (with initial parameters) in *Maze World*. Table VIII presents the same data for *Mirror World* and Table IX averages the data from both worlds. Table X reveals the results of conducting standard one-tailed t-tests on the means of $S$, $t$ and $n$ from these data sets. Throughout this research, differences are accepted as significant at the 95% level but if the difference is also significant at the 99% level this is indicated. The percentages of good and bad runs from this set of experiments are also shown graphically in Fig. 7.

Table XI presents mean and standard deviation data for the experiments using different $k_1$ and $b$ parameters for S3. It includes the performances of S1 and S3 with the original parameters to make comparison easier. Significant differences between these results and the initial S1 results are summarized in Table XII. The weaker learning results are provided in Tables XIII (means and standard deviations) and XIV (significant differences).

### A. Initial Parameters - The Untrained Robots

In *Maze World* S3 shows the best performance in terms of all of the fitness measures and S1 shows the worst, with system S2 second best in each case. In addition, 53% of S3's runs are considered good and none are considered bad, compared with 27% good, 33% bad for S1 and 33% good, 7% bad for S2 (see Fig. 7). S3 is significantly better than S1 at the 99% level for $S$ and $n_s$ and at the 95% level for $t$. Also, S3 is significantly better than S2 and S2 is significantly better than S1 at the 95% level when comparing the mean $n_s$ and $S$ values.

Although S2 out-performs S1, system S3 demonstrates faster and safer results than S2. This indicates that a full implementation of the network is required to elicit a suitable idiotypic response. In S2 there is no global feedback to the system from the communicating antibodies, i.e. the adjusted strength of match values $T_g$ make a difference only within the current iteration and are discarded once the stage 2 competition is finished. However, in S3 the $T_g$ values are incorporated into the updated concentration level through (8) for every antibody in the system. The concentration levels also feed back into the network for the next iteration via (4) and (5), which renders a much more dynamic system. Comparison of S2 and S3 has thus shown that concentration levels have a vital role in mediating the suppressive and stimulatory responses of the idiotypic system, i.e. it is not sufficient just to nominate alternative behaviors on the basis of a fitness metric that is governed only by paratope and idiotope values. The paratope and idiotope comparisons also need to be weighted using a second fitness measure that is non-antigen specific (concentration). Further research into the complex dynamics is clearly needed, but it is apparent that concentration serves to enrich the process by which alternative antibodies are selected. It is possible that S3 is able to discriminate between suitable and inappropriate alternatives in a more efficient manner.

The above observations represent strong statistical evidence that the implementation of a full idiotypic network improves the performance of a reinforcement learning robot, influencing its behavior in a positive manner during the initial learning period. However, analysis of maximum and minimum data reveals that all three systems are capable of fast and safe runs, (the minimum $t$ is between 152 and 156 and the minimum $n_s$ is between 2 and 9 for each system) and all are likely to get into some kind of difficulty (maximum $t$ values are all over 380 and maximum $n_s$ values are all above 45). However, the maximum values for $t$ (485, 406 and 385 for S1 - S3 respectively), the maximum number of stalls (127, 111 and 50) and the lower standard deviations for S3 indicate that the idiotypic robots are somehow protected from executing disastrous runs. In order to investigate this further, the rates of reinforcement success, antibody change and idiotypic difference are examined.

Analysis of the mean rates of reinforcement success reveals an important difference between systems S1 and S3. In S1 the mean is 48%, but this falls to 46% in stall situations. In S2 the mean is the same (50%) in both cases. However, in S3 the overall success rate (49%) rises to 58% when the robot stalls. The difference between S1 and S3 is significant at the 95% level, i.e. S3 produces a significantly better success rate when it stalls than S1. In addition, the difference is significant at the 95% level within S3, i.e. it produces a significantly higher success rate during stalls than overall.

In S3 the mean idiotypic difference rate is 21% rising to 29% during stalls (although the difference is not significant). The mean idiotypic success rate is 20% overall, but increases to 49% when the robot stalls. This represents a significant difference at the 99% level and suggests that, when stalled, the full idiotypic robot is able to choose more successful antibodies by increasing the rate of idiotypic difference. This advocates that a mechanism for recognizing and responding to dangerous situations is inherent in the idiotypic dynamics.

There are no significant differences between S2 and S3 in terms of idiotypic differences and idiotypic success rates. S2's mean idiotypic difference rate is 18% rising to 27% during stalls. The mean idiotypic success rate is 21% in total, increasing to 46% when the robot stalls. This represents a significant difference within the system at the 99% level but is not significantly different to the value of 49% for S3. However, the figures may be misleading because in S2 the process of selecting the stage 1 winner is not influenced by past idiotypic calculations as there is never any global feedback from the network. In S3 the choice of antibody in stage 1 is directly affected by past $T_g$ scores, which means that undetectable idiotypic differences are constantly occurring. Not surprisingly, when the robots are stalled the antibody change rates show significant differences between the systems with S3 demonstrating a rate of 65% rising to 88% when stalled compared with 66% rising to only 78% for S1 and 58% rising to 82% for S2. The change rate for S3 is significantly higher than both the others at the 95% level. However, the observed increase in change rate is significant at the 99% level within each system, which shows that there is a need for rapid antibody change during stall conditions and that all the systems are capable of delivering such changes. This interpretation may be deceptive though, because these results deal with both good and bad runs and collisions lasting only one iteration are also counted as stalls. In order to gain a better understanding it is necessary to consider the sampled long stall data.

Detailed analysis of the long stall sequence data reveals that there is a significant difference at the 99% level for the mean duration of sequences. In S3 the idiotypic robots remain stuck in these sequences for an average of 4.78 iterations before freeing themselves, whereas the non-idiotypic robots in S1 remain trapped on average for 8.53 iterations. In addition, examination of the antibodies used in these trap conditions shows that the mean number of repeated behaviors in S3 is 1.54 compared with 3.42 for S1. This difference is significant at the 95% level. In addition, the antibody change rate during these sequences is 68% for S3 but only 19% for S1. However, there is no significant difference between the means of the numbers of long stall sequences. These observations support the view that idiotypic robots are able to out-perform their non-idiotypic counterparts by freeing themselves from stalls more quickly. In addition, it suggests that their rapid escape is accomplished by an ability to switch behaviors at a higher rate. This provides good evidence to support hypotheses $H_2$ and $H_3$ and is further substantiated by analysis of the idiotypic differences in S3 for this sub-set of the data. On average 72% of S3's long stall sequences are terminated (i.e. the robot escapes) when an idiotypic difference occurs. Moreover in 63% of the long sequences the idiotypic difference generates an untried antibody that proves successful when the stage 1 matching process is still suggesting the use of antibodies that have already failed. This analysis helps to explain the large differences in the standard deviations between S1 and S3 and fits in with the other observations. All the systems are capable of performing well, but when stall problems occur it seems that S3 is able to resolve them more rapidly, which means that it is not inclined to produce disastrous runs.

### B. Initial Parameters - The Trained Robots

In *Mirror World*, where the initial paratope matrix is taken as the output from *Maze World*, each system improves on the mean $n_s$, $t$, and $S$ from its *Maze World* trials, which demonstrates that all three systems allow a degree of learning to take place. In addition, standard deviations are generally lower as there are far fewer bad performances and an improvement in $S$ is demonstrated on 77%, 70% and 80% of runs for S1 - S3 respectively. The percentage of good runs also increases from the first environment to 60%, 57% and 80% for the three systems. However, although S3 produces no unacceptable runs, S1 and S2 still under-perform for approximately 7%. S3 again shows the best performance in terms of all of the metrics and consistently demonstrates a lower standard deviation than S1. It is also significantly better than S1 at the 99% level for $S$ and $n_s$. In addition, S3 shows that it is significantly better than S2 at the 95% level for all the criteria. S2 produces the second best results on all counts except for time to complete the task, but its performance is not significantly better than S1.

The observations and analysis of the *Mirror World* data suggest that the robots are less likely to get into difficulties when using a paratope matrix from a previous run, because they have developed good obstacle avoidance strategies from their earlier experiences. This means that stalls tend to happen less frequently and run times are generally faster. However, significant differences are still apparent between S1 and S3 and S2 and S3, suggesting that the full idiotypic network still has an important role to play in assisting robots to escape from traps after initial learning has taken place. This supports the hypothesis $H_1$, that idiotypic systems permit a degree of de-coupling from an engineered learning system, since it alludes to the fact that the idiotypic network is still able to influence a reinforcement system positively, even after the robot has had ample time to complete the learning process.

*C. Initial Parameters - The Combined Results*

Analysis of the combined results, i.e. means of the two $n_s$, $t$, and $S$ values from each run, shows that S3 is significantly better than S1 and S2 for all three fitness measures, even task time. These differences are at the 99% level for $S$ and $n_s$ and at the 95% level for $t$. The effect of combining the results in this way is to smooth out the data, reducing the standard deviations, which allows good comparison. From Table X it is readily apparent that S3 is superior both to S2 and S1. In addition, the percentages of good and bad runs reflect the incremental nature of the systems, with S3 performing well in 67% of all runs and badly in none, S2 running well in 57% and badly in 3% and S1 achieving a good run in 30% and a bad run in 20%, see Fig. 3. This analysis provides good evidence that the full idiotypic network can significantly improve robot performance during longer tasks, i.e. through tasks that include a learning phase and a mature phase, where a stable paratope matrix has developed.

*D. Varying Parameters*

Table XI shows that S3 achieves a remarkably similar performance to its first trial when different parameters are used. This is especially true for the mean and standard deviation of the score and number of stalls. As in the first trial these are both significantly better than S1's performance at the 99% level.

Moreover, S3 improves on its original mean time of 237 seconds, yielding mean completion times of 215 and 225 seconds for $b = 100$ and $b = 60$ respectively. These increased levels of performance permit significant differences between S3 and S1 at the 99% level for time, rather than at the 95% level in the original data. This provides even stronger evidence to support the case for the full idiotypic advantage and demonstrates that there is a degree of flexibility within the $k_1$ and $b$ parameters. These additional sets of results may also indicate that 0.625 is slightly too high to be optimum for $k_1$ in this region of $b$ and with this reinforcement scheme.

*E. The Weaker Learning Strategy*

Comparison of Tables VII and XIII shows that the means and standard deviations of $S$, $t$ and $n_s$ increase for both systems when the weaker strategy is implemented. In addition, the number of good scores reduces and the number of bad scores increases, reflecting the fact that obstacle avoidance is more difficult to achieve. Within the weak learning experiments, the differences between mean $S$, $t$ and $n_s$ for S1 and S3 is significant at the 99% level. This is further evidence in support of the implementation of a full idiotypic network to accompany reinforcement learning, but the real value of this experiment lies in comparing S3's weak learning performance with S1's strong learning performance. S3 achieves a mean ($n_s$, $t$, $S$) of (27, 273, 257) with weak learning compared with (45, 277, 342) for S1 using strong learning. The difference between the $n_s$ values is significant at the 99% level and the difference between the $S$ values is significant at the 95% level. This means that robots implementing the full idiotypic network and poor learning are performing as well as (possibly better than) robots with good learning but no idiotypic selection, which suggests that a full network may be able to offer a degree of compensation for poor learning. This supports hypothesis $H_1$ and shows that the idiotypic robots may have been implementing more creative solutions to the problem.

VIII. CONCLUSION AND FUTURE AIMS

*A. Conclusion*

A computational method for simulating idiotypic effects is developed, based on Farmer's popular model of Jerne's idiotypic network. The scheme is incorporated into a reinforcement learning (RL) architecture and compares antibody idiotopes and paratopes in order to determine inter-antibody suppression and stimulation levels. The architecture is fully described and incrementally implemented with virtual robots that perform a color-tracking task in order to test three hypotheses $H_1 - H_3$. $H_1$ asserts that idiotypic systems allow a degree of detachment from reinforcement learning, $H_2$ proposes that they reduce premature convergence and $H_3$ postulates that they allow escape from repeated behavior patterns.

The use of the full idiotypic network (S3) produces significantly better results than partial implementations that use RL only (S1) and a simplified network without global feedback (S2), thus highlighting the benefits of introducing greater idiotypic complexity. The faster and safer performance of S3 is chiefly attributed to its ability to recover from stall situations much more rapidly than the other systems, which is thought to be a direct result of idiotypic activity. Indeed this paper provides evidence to suggest that during a sequence of stalls, S3 is capable of increasing the rate of antibody change autonomously so that repeated behaviors are discarded in favor of suitably selected alternatives. This may be a result of the system's ability to raise the rate of idiotypic communication during a stall, so that a much higher reinforcement success rate is achieved, implying that idiotypic networks have an inherent mechanism for detecting and responding to trap situations. These results confirm the likelihood of hypothesis $H_2$, as an increased rate of antibody change implies a much less greedy strategy. In addition, this paper supplies evidence that during stall sequences, the idiotypic process tends to generate previously untried successful antibodies whilst the antigen matching process recommends repeated failures. This is direct evidence for the support of $H_2$ and also upholds $H_3$. However, since repeated loops of behavior may also occur in non-stall situations, further tests that isolate recurring behavior patterns are recommended to test $H_3$ further.

The simplified idiotypic system S2 is believed to have under performed in comparison to S3 because of the lack of global feedback from the idiotypic network to the antibody concentrations and vice versa. Its inferior attainment demonstrates that concentration and feedback are extremely important components of an idiotypic system, possibly providing an additional memory feature that allows discrimination between suitable and inappropriate alternatives in a more efficient manner.

Evidence to support $H_1$ is provided by comparing performance of the systems after training, where S3 still proves superior to S1. This shows that the network retains its influence over the system once learning has taken place, i.e. that there is a sense of de-coupling from the reinforcement strategy. Furthermore, when S3 is implemented with a weaker learning strategy, its performance is still significantly better than with S1 using stronger learning. This clearly suggests that a full idiotypic network permits robots greater scope for creating solutions to the task as they are able to assert a degree of independence over behaviors prescribed by the engineered reinforcement signals.

### B. Future Aims

It may be argued that using a hand designed idiotope is equivalent to providing the robot with *a priori* information about the behaviors, since it effectively shows which are of similar type. Indeed, the idiotypic selection algorithm may be regarded as somewhat redundant when using a contrived matrix such as this because it is readily apparent which antibodies are of similar type and which are different. The next step to this research is therefore to investigate whether similar results can be obtained when an initially random variable idiotope is used. The variable matrix would develop by incrementing antibody-antigen combinations that produce a high rate of negative reinforcement learning scores as in [15]. Once a self regulating and variable idiotope is in place a meta-dynamic system with mutation may be designed and concentration levels can be used to determine which antibodies are retained and which die. Future research will therefore focus on developing means for creating new antibodies, testing and mutating them.

In addition, further investigation into the complex dynamics of the full idiotypic network is the logical extension to this work. In particular, more extensive research into the relationships between the parameters $k_1$, $k_2$ and $b$ will be conducted with testing taking place using a wide variety of environments, reinforcement schemes, problems, robots and antibody selection mechanisms. The effect of the idiotope matrix upon these parameters will also be studied by testing different fixed matrices and several variable schemes.

Moreover, since it is always extremely difficult to know whether simulation results generalize to the real world, these systems will also be trialed using real robots that attempt to solve similar problems in dynamically changing environments. It is possible that an idiotypic network may bring even more advantage to a real world system, since it is less predictable and should therefore require a less pre-determined method of behavior selection. However, prior to this it is necessary to develop a method that can provide reasonably good starting paratopes, allowing close to zero stalls for the real robots. A major part of extending this work will therefore involve integrating the AIS system with a genetic algorithm that will run in highly accelerated simulations, evolving a strong set of base rules to initialize the real robot.

REFERENCES

[1] N. K. Jerne, "Towards a network theory of the immune system", *Ann. Immunol. (Inst Pasteur)*, 125 C, pp. 373-389, January 1974

[2] J. D. Farmer, N. H. Packard, A. S. Perelson, "The immune system, adaptation, and machine learning", *Physica*, D, Vol. 2, Issue 1-3, pp. 187-204, October – November 1986

[3] F. M. Burnet, *The clonal selection theory of acquired immunity*, Cambridge University Press, 1959

[4] L. N. de Castro, J. Timmis, *Artificial immune systems: A new computational intelligence approach*, London, Springer-Verlag, 2002

[5] Y. Watanabe, A. Ishiguro, Y. Shirai, Y. Uchikawa, "Emergent construction of behavior arbitration mechanism based on the immune system", in *Proceedings of the 1998 IEEE International Conference on Evolutionary Computation, (ICEC)*, pp. 481-486, May 1998

[6] T. Kondo, A. Ishiguro, Y. Watanabe, Y. Shirai, Y. Uchikawa, "Evolutionary construction of an immune network-based behavior arbitration mechanism for autonomous mobile robots", *Electrical Engineering in Japan*, Vol. 123, No. 3, pp. 1-10, December 1998

[7] P. A. Vargas, L. N. de Castro, R. Michelan, "An immune learning classifier network for autonomous navigation", *Lecture Notes Computer Science*, 2787, pp. 69-80, September 2003

[8] P. A. Vargas, L. N. de Castro, F. J. Von Zuben "Mapping artificial immune systems into learning classifier systems"*, Lecture Notes in Artificial Intelligence*, 2661, pp. 163-186, (IWLCS September 2002), Lanzi, P. L. *et al*. (eds.), 2003

[9] G. C. Luh, W. W. Liu, "Reactive immune network based mobile robot navigation", *Lecture Notes Computer Science*, 3239, pp. 119-132, September 2004

[10] J. Suzuki, Y. Yamamoto, "Building an artificial immune network for decentralized policy negotiation in a communication end system: Open webserver/iNexus study", in *Proc. of the 4$^{th}$ World Conference on Systemics, Cybernetics and Informatics*, (SCI 2000), Orlando, FL, USA, July 2000

[11] D. Chowdhury, "Immune network: an example of complex adaptive systems", *Artificial Immune Systems & Their Applications,* Dasgupta, D. (ed.), Springer, pp. 89-104, 1999

[12] R. Michelan, F. J. Von Zuben, "Decentralized control system for autonomous navigation based on an evolved artificial immune network", in *Proceedings of the 2002 Congress on Evolutionary Computation*, Vol. 2, pp. 1021-1026, (CEC2002), Honolulu, Hawaii, May 12-17 2002

[13] M. Krautmacher, W. Dilger, "AIS based robot navigation in a rescue scenario", *Lecture Notes Computer Science*, 3239, pp. 106-118, September 2004

[14] D. Floreano, F. Mondada, "Evolution of homing navigation in a real mobile robot", *IEEE Transactions on Systems, Man and Cybernetics – Part B Cybernetics*, 26 (3), pp. 396-407, June 1996

[15] A. Ishiguro, T. Kondo, Y. Watanabe, Y. Uchikawa, "A reinforcement learning method for dynamic behavior arbitration of autonomous mobile robots based on the immunological information processing mechanisms", *Trans. IEE of Japan*, 117-C, No. 1, pp. 42-49, January 1997

[16] G. Parker, "Co-evolving model parameters for anytime learning in evolutionary robotics", *Robots and Autonomous Systems*, 33, pp. 13-30, October 2000

[17] V. Gullapalli, "Skillful control under uncertainty via direct reinforcement learning", *Robots and Autonomous Systems*, 15, pp. 237-246, August 1995

[18] M. J. Matarić, "Reinforcement learning in the multi-robot domain", *Autonomous Robots*, 4 (1), pp. 73-83, March 1997

[19] R. T. Vaughan, B. P. Gerkey, A. Howard, "The Player/Stage project: Tools for multi-robot and distributed sensor systems", in *Proceedings of the International Conference Advanced Robotics (ICAR 2003)*, pp. 317-323, Coimbra, Portugal, June 30 – July 3, 2003

[20] P. F. Stadler, P. Schuster, A. S. Perelson, "Immune networks modeled by replicator equations", *J. Math. Biol.*, 33 (2), pp. 111-137, 1994

[21] S. Cayzer, U. Aickelin, "A recommender system based on idiotypic artificial immune networks", *Journal of Mathematical Modelling and Algorithms*, 4 (2), pp. 181-198, 2005

[22] R. A. Brooks, "A robust layered control system for a mobile robot", *IEEE Journal of Robotics and Automation*, RA-2 (1), pp. 14-23, March 1986

[23] S. Sathyanath, F. Sahin, "AISIMAM – An AIS based intelligent multi agent model ands its application to a mine detection problem", in *Proceedings of the ICARIS 2002 1st International Conference on Artificial Immune Systems*, Canterbury, UK, September 9 – 11, 2002

[24] S. Sathyanath, F. Sahin, "Application of Artificial Immune System based intelligent multi agent model to a mine detection problem", in *Proceedings of the SMC 2002, IEEE International Conference on Systems, Man, and Cybernetics*, Vol. 3, Tunisia, October 2002

[25]  W. J. Opp, F. Sahin, "An Artificial Immune System Approach to mobile sensor networks and mine detection", in *Proceedings of the SMC 2004, IEEE International Conference on Systems, Man, and Cybernetics*, Vol. 1, pp. 947-952, October 10-13, 2004

BIOGRAPHIES

**Amanda M. Whitbrook** received the B.Sc. (Hons) degree in Mathematics and Physics from The Nottingham Trent University, U.K. in 1993, the M.Sc. degree in Management of Information Technology from the University of Nottingham, U.K. in 2005 and the Ph.D. degree in numerical analysis from The Nottingham Trent University, U.K. in 1998.

She has previously been employed as a programmer and data analyst and as an information systems designer. She currently holds the position of Research Fellow within the Automated Scheduling, Optimization and Planning (ASAP) Research Group in the School of Computer Science at the University of Nottingham, U.K. Her research interests include robotics, artificial intelligence, adaptive learning, evolutionary algorithms and artificial immune systems.

**Uwe Aickelin** (M'06) received a Management Science degree from the University of Mannheim, Germany, in 1996 and a European Master and Ph.D. in Management Science from the University of Wales, Swansea, U.K., in 1996 and 1999, respectively.

Immediately following his Ph.D., he joined the University of the West of England in Bristol, U.K. where he worked for three years in the Mathematics Department as a lecturer in Operational Research. In 2002, he accepted a lectureship in Computer Science at the University of Bradford, U.K., mainly focusing on computer security. Since 2003 he has worked for the University of Nottingham, U.K., in the Automated Scheduling, Optimization and Planning (ASAP) Research Group in the School of Computer Science where he is now a Reader in Computer Science and Director of the Inter-disciplinary Optimization Laboratory. He currently holds an EPSRC Advanced Fellowship focusing on artificial immune systems, anomaly detection and mathematical modeling. In total, he has been awarded over £2 million EPSRC research funding as Principal Investigator (including an Adventure Grant and two IDEAS Factory projects) on topics including artificial immune systems, danger theory, computer security, robotics and agent based simulation.

Dr. Aickelin is an Associate Editor of the IEEE *Transactions on Evolutionary Computation*, the Assistant Editor of the *Journal of the Operational Research Society* and an Editorial Board member of *Evolutionary Intelligence*.

**Jonathan M. Garibaldi** received the B.Sc. (Hons) degree in Physics from Bristol University, U.K., and the M.Sc. degree in Intelligent Systems and the Ph.D. degree in uncertainty handling in immediate neonatal assessment from the University of Plymouth, U.K., in 1984, 1990, and 1997, respectively.

He is currently an Associate Professor within the Automated Scheduling, Optimization and Planning (ASAP) Research Group in the School of Computer Science at the University of Nottingham, U.K. The ASAP research group tackles a wide range of decision making problems with particular emphasis on heuristic and meta-heuristic approaches to combinatorial optimization. He has published over 40 papers on fuzzy expert systems and fuzzy modeling, including three book chapters, and has edited two books. His main research interests are modeling uncertainty in human reasoning and in the intelligent analysis of large, multimodal, noisy data sets. He has created and implemented fuzzy expert systems, and developed methods for fuzzy model optimization.

FIGURE CAPTIONS

Fig. 1. Showing antibody paratope and idiotope regions and inter-antibody stimulation and suppression [4].

Fig. 2. The Maze World used for conducting the door tracking experiments with untrained robots.

Fig. 3. Mean score versus $k_1$ for S2 and S3 with $b$ values of 8, 80 and 800.

Fig. 4. Idiotypic difference rate versus $k_1$ for S2 and S3 with $b$ values of 8, 80 and 800.

Fig. 5. Mean score versus $b$ for S3 with $k_1$ between 0.45 and 0.65

Fig. 6. Mean score versus $k_1$ between 0.45 and 0.65 for S3 in the region $40 \leq b \leq 160$

Fig. 7. Histogram of good and bad runs for the three systems, S1 – S3.